\documentclass{article}

\usepackage[nonatbib]{sty_2023}

\usepackage[utf8]{inputenc} 
\usepackage[T1]{fontenc}    
\usepackage{hyperref}       
\usepackage{url}            
\usepackage{booktabs}       
\usepackage{amsfonts}       
\usepackage{nicefrac}       
\usepackage{microtype}      
\usepackage{xcolor}         
\usepackage{algorithm}
\usepackage[noend]{algpseudocode}
\usepackage{amsmath}
\title{FedSplitX: Federated Split Learning for Computationally-Constrained Heterogeneous Clients}
\usepackage{graphicx}
\usepackage{multirow}
\usepackage{parskip}
\usepackage{bm}
\usepackage{subcaption}
\usepackage{multirow}
\usepackage{siunitx}

\author{%
  Jiyun ~Shin \\
  KAIST \\
  \texttt{gyoon814@kaist.ac.kr} \\
  \And
  Jinhyun ~Ahn \\
  Myongji ~University \\
  \texttt{wlsgus3396@mju.ac.kr} \\
  \AND
    Honggu ~Kang \\
  KAIST\\
  \texttt{khg13@kaist.ac.kr} \\
  \And
  Joonhyuk ~Kang \\
  KAIST \\
  \texttt{jkang@kaist.ac.kr} \\
}

\begin{document}

\maketitle

\begin{abstract}

Foundation models (FMs) have demonstrated remarkable performance in machine learning but demand extensive training data and computational resources. Federated learning (FL) addresses the challenges posed by FMs, especially related to data privacy and computational burdens. However, FL on FMs faces challenges in situations with heterogeneous clients possessing varying computing capabilities, as clients with limited capabilities may struggle to train the computationally intensive FMs. To address these challenges, we propose \textit{FedSplitX}, a novel FL framework that tackles system heterogeneity. FedSplitX splits a large model into client-side and server-side components at multiple partition points to accommodate diverse client capabilities. This approach enables clients to collaborate while leveraging the server's computational power, leading to improved model performance compared to baselines that limit model size to meet the requirement of the poorest client. Furthermore, FedSplitX incorporates auxiliary networks at each partition point to reduce communication costs and delays while enhancing model performance. Our experiments demonstrate that FedSplitX effectively utilizes server capabilities to train large models, outperforming baseline approaches. 

\end{abstract}

\section{Introduction}
Foundation models (FMs), that is large machine learning model, show remarkable performance in building powerful machine learning systems. However, FMs require extensive amounts of training data and computational power \cite{bommasani2021opportunities}. Acquiring such massive training data poses challenges related to data privacy, legal constraints \cite{villalobos2022will}, and computational burdens \cite{touvron2023llama},\cite{brown2020language}. Federated learning (FL) \cite{mcmahan2017communication} is a promising distributed machine learning framework that can solve the such challenges of FMs \cite{zhuang2023foundation}. The local clients in FL with local training data collaboratively train a global model by aggregating locally-updated model parameters without sharing private local data. In FL, there can be various clients with different computing and communication capabilities. Among them, some clients with limited resources may encounter difficulties in training the global model. In such cases, FL can be performed by reducing the size of the global model or by allowing only clients with sufficient resources to participate in learning. However, this approach can lead to a degradation in the performance of the global model. Some previous works studies to tackle these system-heterogeneous clients by scaling down the global model into sub-models \cite{horvath2021fjord, diao2021heterofl,kang2023nefl,kim2023depthfl}, however, reduced model size degrades the performance due to its limited model capacity.

To address these resource-constrained challenges, split learning (SL) proposes to train a global model employing server \cite{vepakomma2018split, gupta2018distributed} without scaling down a model and accessing raw data from clients. It splits the model by a partition point (i.e., a specific layer) and let clients train partial model up to the specific layer. The clients transmit output at the specific layer to a server known as the smashed data. Then, a server trains the remaining layers. Split federated learning (SFL) \cite{thapa2022splitfed} proposed an algorithm that combines SL and FL to reduce the training time. However, since SL and SFL split a model with a single partition point, the partition point should satisfy the requirements for the clients of the poorest computing capabilities. Then, clients with sufficient capabilities may not fully utilize their potential, arising computing burden to server. It will be more critical for training a large model.

In this paper, we propose a novel FL framework, \textbf{FedSplitX} to tackle system heterogeneity with limited computing capabilities. In FedSplitX, a large model is partitioned into client-side model and server-side model with multiple partition points to meet the heterogeneous capabilities. By employing multiple partitioning points, multiple pairs of client-side and server-side models are created depending on the different partition points. We also introduce auxiliary networks for the different partition points. These auxiliary networks enable local-loss-based learning of client-side model, reducing the communication costs and delays. Furthermore, the collaborative loss from auxiliary networks enhances performance. For local model optimization, our framework aggregates parameters of client-side model and server-side model separately by aggregation scheme from \cite{diao2021heterofl} method, which computes the average only for updated parameters. 
Our experiments show that FedSplitX can utilize the abilities of the server to train large model that require more computing power than client capabilities, resulting in better performance compared to baseline, where the global model size is limited to match the client's computing power. In particular, we verify that the collaborative loss with auxiliary networks at each partition point can effectively improve performance.

Our contributions are summarized as follows:
\begin{itemize}
    \item We propose FedSplitX, a framework that allows heterogeneous clients to cooperate by utilizing the server's capabilities.
    \item We demonstrate that FedSplitX that trains a large model aided by a server, outperforms baselines that fits the requirements of all the clients
    \item We show that FedSplitX splits the model with multiple partition points, allowing clients to fully utilize their own capabilities while reducing the computational load on the server.
\end{itemize}

\begin{figure}[t]
    \centering
    \begin{subfigure}{0.8\textwidth}
        \centering
         \noindent
        \makebox[\textwidth]{
        \includegraphics[width=1.25\textwidth]{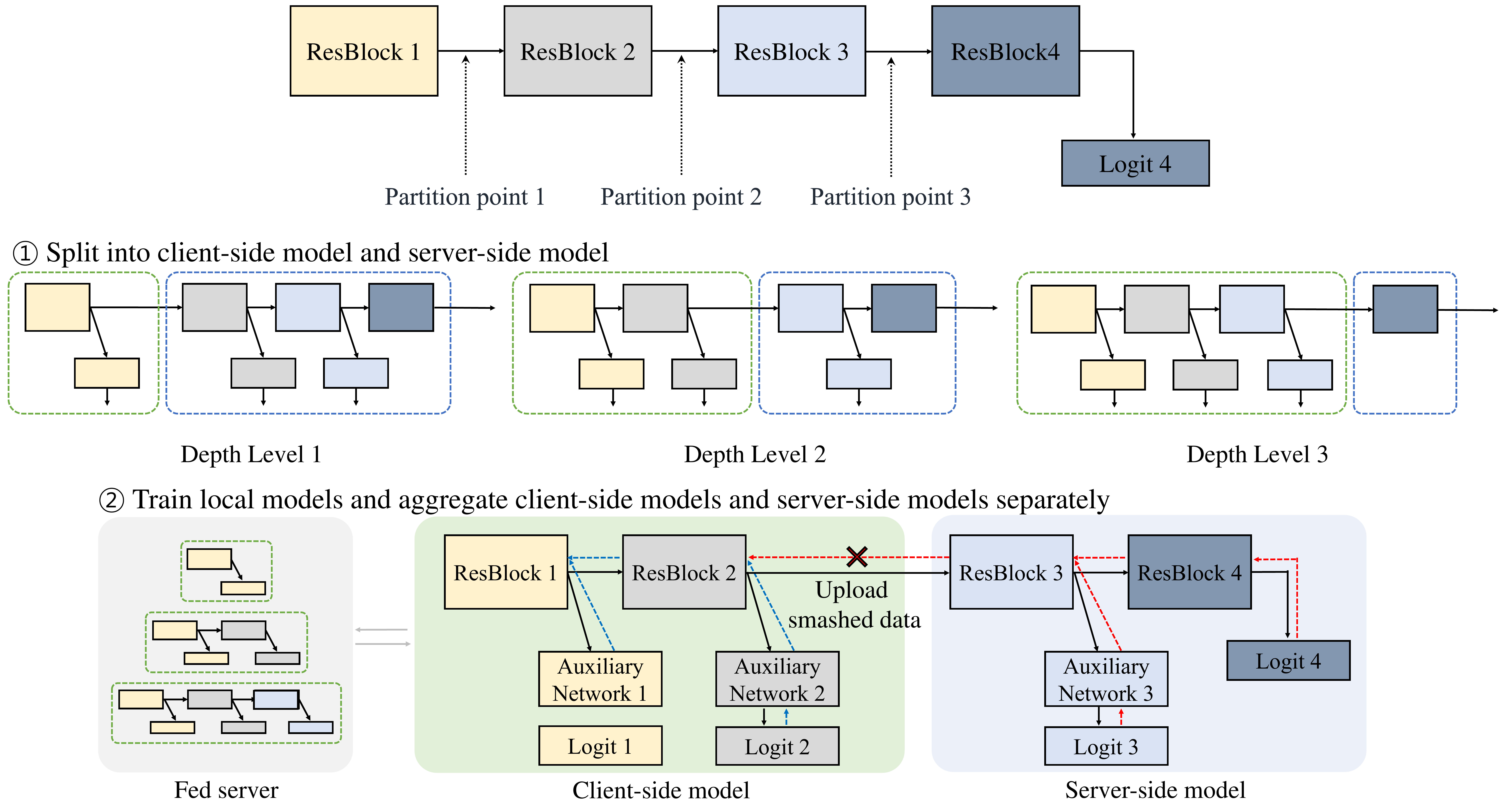}
        }
        \label{fig:steps}
    \end{subfigure}
 
    \caption{Framework of FedSplitX in client-system with $M = 3$ depth-levels. (top) The overall structure of the model to be trained and the partition points based on depth-level, (middle) Client-side model and server-side model partitioned based on partition point according to depth-level , (bottom) Training process after split operation. }
    \label{fig:pruning}
\end{figure}

\newpage
\section{FedSplitX: Federated Split Learning in Heterogeneous Client System} 
\label{gen_inst}

We suppose that $K$ heterogeneous clients participate in FL. Each client has its local training dataset of size $N_k$, denoted by $\mathcal{D}_k = \{(\mathbf{x}_i, y_i)\}_{i=1}^{N_k}$ where $k \in \{1,\ldots,K\}$. The clients are clustered into $M$ clusters based on their computing capabilities. In other words, we consider that we have a total of $M$ types of heterogeneous clients. We label each cluster in ascending order based on the clients' capabilities within that cluster, and this labeling is defined as the \textit{depth-level}. A depth-level of a client $k$ who involves in a cluster $m$ is designated as $d_k=m$ where $m\in\{1,\dots,M\}$. Note that clients in cluster with depth-level $M$ have the highest computing capability and clients in the cluster with depth-level $1$ have the lowest computing capability. For every round of FedSplitX, each client trains its own split model and transmits smashed data to main server. The main server uses this smashed data to train remaining partitioned model. During server training, clients independently train their split model using \textit{local-loss}. The clients upload parameters of trained models to the fed server for aggregation, and the server-trained models are aggregated on the main server. The details are described in Algorithm \ref{alg1}.

\begin{algorithm}[t!] 
\caption{Federated Split Learning with Heterogeneous Clients (FedSplitX)}
\textbf{Input:} Dataset $\mathcal{D}_k$ on client $k$, distributed $K$ local clients, the fraction $C$ of clients per communication round, {the number} local epochs $E$, the local minibatch size $B$, the learning rate $\eta$, the global client-side model parameterized $\mathbf{w}^{c}$, the global server-side model parameterized $\mathbf{w}^{s}$, the computing level $d_k \in \{1,2,\ldots,M\}$.
\begin{algorithmic}[1] 
    \State \textbf{Server executes:}
    \For {round $t= 0,1, \ldots, T-1$}
        \State $P_t \leftarrow $ Random Clients
        \For{each client $k \in P_t$, and in parallel}
            \State Download the client-side model from fed server $\mathbf{w}^c_{k} \leftarrow  \mathbf{w}^s[:d_k]$
            \State Extract server-side model from global server-side model $\mathbf{w}^s_{k} \leftarrow  \mathbf{w}^s[d_k:]$
            \State $(\mathbf{s}_{k}^{c}, \mathbf{y}_k) \leftarrow $ \textbf{GetSmashedData}$(\mathbf{w}^c_k, \mathcal{D}_k)$
            \State $\mathbf{w}^c_{k} \leftarrow $ \textbf{Client\_Update}$( \mathbf{w}^c_{k})$
            \For{local epoch $e$ = 1,2,\ldots,E}
                \State Forward propagation with $\mathbf{s}_{k}^{c}$ on $\mathbf{w}^{s}_{k}$
                \State $\mathbf{w}^{s}_{k} \leftarrow \mathbf{w}^{s}_{k} - \eta \nabla F^s_{k}(\mathbf{w}^{s}_{k}, \mathbf{a}_{[d_k+1:M]})$
            \EndFor
        \EndFor
        \State $\mathbf{w}_{k}^{c} \leftarrow$  \textbf{heteroavg} $\left(\{\mathbf{w}_{k}^{c}\}_{k=1}^{K}\right)$ , $\mathbf{w}_{k}^{s} \leftarrow$  \textbf{heteroavg}$\left(\{\mathbf{w}_{k}^{s}\}_{k=1}^{K}\right)$
    \EndFor
\medskip
\State\textbf{Client\_Update $(\mathbf{w}^c_{k})$:}
\For {local epoch $e = 1,2, \ldots, E$}
    \State Forward propagation with  $\mathcal{D}_k$ on $\mathbf{w}^c_{k}$
    \State $\mathbf{w}^c_{k} \leftarrow \mathbf{w}^c_{k} - \eta \nabla F^c_{k}(\mathbf{w}^c_{k}, \mathbf{a}_{[1:d_k]})$
\EndFor
\medskip
\State\textbf{GetSmashedData}$(\mathbf{w}^c_k, \mathcal{D}_k)$:
    \State  $\mathbf{s}_{k}^{c}, \; \mathbf{y}_{k} \leftarrow \{ \mathbf{w}_{k}^{c} \left( \mathbf{x}_{i}  \right) \} , \{ y_{i} \},$ where\;\;  $(\mathbf{x}_{i} ,y_{i}) \in D_{k}$

\end{algorithmic}
\label{alg1}
\end{algorithm}
\subsection{Heterogeneous Split Models with Auxiliary Networks}

We split the global model to address FL problems in resource-constrained client system. The full model, denoted as $\mathbf{w} = [\mathbf{w}^c, \mathbf{w}^s]$ is split into two parts: a client-side model, denoted as $\mathbf{w}^c$, and a server-side model, denoted as $\mathbf{w}^s$.

The partition point of a client is determined based on the computation capabilities of the client. In heterogeneous client system with $M$ depth-levels, we can split the entire model using $M$ different partition points, each associated with a specific depth-level. Consequently, we can obtain $M$ pairs of client-side and server-side models, each with different partition points. The client-side model and server-side model for depth-level $m$ are represented as $\mathbf{w}^c_{m} = \mathbf{w}^c[:m]$ and $\mathbf{w}_m^{s} = \mathbf{w}^s[m:]$, where $m\in\{1,\dots,M\}$. We define $\mathbf{w}^c = \mathbf{w}^c_{M}$ as the largest client-side model and $\mathbf{w}^s = \mathbf{w}^s_{1}$ as the largest server-side model.

We consider auxiliary networks for local-loss based learning and collaborative loss, which is detailed in Section \ref{local_train}. Auxiliary networks are connected at all the partition points that exist in both client-side model and server-side model.
For example, a client system with a total of $M$ depth-levels, there are $M$ auxiliary networks $\mathbf{a}_{[1:M]}$. A client $k$ of depth-level $d_k=m$ has $m$ out of the $M$ auxiliary networks $\mathbf{a}_{[1:m]}$ that are connected to the client-side model, while the remaining $M - m$ auxiliary networks $\mathbf{a}_{[m+1:M]}$ are connected to the server-side model.

\subsection{Local Model training with Auxiliary Networks}
\label{local_train}

For local training, each client downloads parameters of local client-side model from the fed server. A client $k$ with depth-level $d_k=m$ trains the client-side model $\mathbf{w}_m^{c} = \mathbf{w}^c[:m]$ and performs forward propagation for their local data samples in parallel. The client $k$ obtains smashed data $\mathbf{s}_{k}^{c}$ at the end of the client-side model for all data samples, then uploads smashed data $\mathbf{s}_{k}^{c}$ to the main server. Then, the main server performs forward propagation based on the smashed data, calculates the loss, and performs backward propagation based on this loss.

In SFL, there is a communication delay of transmitting the gradients that are sent from the server-side model to the client-side model for an backward propagation. Additionally, the transmission of the gradient at each local epoch incurs significant communication costs. We aim to reduce these delay and communication cost by utilizing the last auxiliary network $\mathbf{a}_{[m]}$ on the client-side, which is attached to the end of the client-side model. Now, clients can computes local-loss and performs backward propagation using $\mathbf{a}_{[m]}$ without waiting for gradients from the server. Simultaneously, the main server updates the server-side model using smashed data $\mathbf{s}_{k}^{c}$.

To address model heterogeneity, we obtain \textit{collaborative loss} with logits from the auxiliary networks. The collaborative loss ensures that each client-side model operates effectively as an independent model, regardless of the partition point chosen. 
In the scenario where depth-level of client $k$ is $m$, on the client-side, there exist $m$ intermediate logits $\mathbf{l}_i^{k},\; i = 1,2,\ldots,m$ obtained from each $m$ auxiliary networks. We sum the loss between intermediate logits $\mathbf{l}_i^{k},\; i = 1,2,\ldots,m$ of client and the ground truth label $y^k$ to obtain the loss function for the client-side model. Client-side models $f_c$ are trained to minimize their respective loss functions $F_c(\cdot)$ as follows:
 \vspace{-0.2in}

\begin{gather*}
\min_{\mathbf{w}^c,\mathbf{a}} F_c(\mathbf{w^c}) = \min_{\mathbf{w}^c,\mathbf{a}}\frac{1}{K}\sum_{k=1}^{K} F_{k}^c(\mathbf{w}^{c}_{d_k}, \mathbf{a}_{[1:d_k]}),\\[0.8pt]
\text{where }
 F_{k}^c(\mathbf{w}^{c}_{d_k}, \mathbf{a}_{[1:d_k]}) = \frac{1}{|D_k|}\sum_{j=1}^{|D_k|}\sum_{i=1}^{d_k}\ell_c\left(\mathbf{l}_{i,j}^{k},y_j^k\right),\;
\mathbf{l}_{i}^{k} = \mathbf{a}_{[i]}\left(f_k^{c}[:i](\mathbf{w}^{c}_{d_k};\mathbf{x})\right)
\end{gather*}
On server-side, intermediate logits can be obtained from the $M-m$ auxiliary networks connected to the server-side model, and logits from the model's output can also be obtained. We similarly use the sum of losses between these logits and ground truth on the server-side as the loss function for the server-side model. Server-side models $f_s$ are trained to minimize server-side loss functions $F_s(\cdot)$ as follows:
 \vspace{-0.1in}
\begin{gather*}
\min_{\mathbf{w}^s,\mathbf{a}} F_s(\mathbf{w^s}) = \min_{\mathbf{w}^s,\mathbf{a}}\frac{1}{K}\sum_{k=1}^{K} F^{s}_{k}(\mathbf{w}^{s}_{d_k}, \mathbf{a}_{[{d_k}+1:M]}),\\
\text{where }
F_{k}^s(\mathbf{w}^{s}_{d_k}, \mathbf{a}_{[{d_k}+1:M]}) = \frac{1}{|D_k|}\sum_{j=1}^{|D_k|}\left(\sum_{i={d_k}+1}^{M+1}\ell_s(\mathbf{l}_{i,j}^k,y_j^k)\right),\\[0.8pt]
\mathbf{l}_{i}^{k} = \mathbf{a}_{[i]}\left(f_s^{k}[:i](\mathbf{w}^{s}_{d_k};\mathbf{s}^c_{k})\right),\;\;i \in [{d_k}+1,\ldots,M], \text{ and } \mathbf{l}_{M+1}^{k} = f^s_{k} (\mathbf{w}^{s}_{d_k};\mathbf{s}^c_{k}).
\end{gather*}
\subsection{Local Model Aggregation} 
FedSplitX aggregates the client-side model and server-side model separately. 
The aggregation of server-side model parameters is performed at the main server, while the aggregation of client-side model parameters takes place at the fed server. The fed server is a server dedicated to aggregating client-side model parameters. It receives parameters uploaded by clients, performs parameter aggregation.
Heterogeneity of model leads to heterogeneity in the model updates and, hence, we need to account for that in the global aggregation as follows \textit{heteroavg} \cite{diao2021heterofl, horvath2021fjord}: 
\begin{gather*}
\mathbf{w}^{c}_{1} = \frac{1}{K}\sum_{i=1}^{K}\mathbf{w}^{c}_{i,1}, \;\;\mathbf{w}^{c}_{m}\backslash \mathbf{w}^{c}_{m-1} = \frac{1}{K - K_{1:m}}\sum_{d_i \ge m}\mathbf{w}^{c}_{i,m} \backslash \mathbf{w}^{c}_{i,m-1} \,\;\; m = 2,\ldots,M\\
\mathbf{w}^{c} = \mathbf{w}^{c}_1 \cup (\mathbf{w}^{c}_2  \backslash  \mathbf{w}^{c}_1) \cup  \dots \cup (\mathbf{w}^{c}_M \backslash \mathbf{w}^{c}_{M-1}).
\end{gather*}
At the same time, the main server averages the parameters of the server-side models in the same way.

\section{Experiments}
\label{headings}
In this section, we present the evaluation of our proposed FedSplitX compared to the baselines. All experiments are trained for $1000$ global rounds with $1$ local epoch per round. We considered $K=50$ clients, where $10\%$ of the clients are randomly selected to participate in the training during each global round. We set batch size to 64 and used Stochastic gradient \cite{ruder2016overview}. For inference in EXC, FjORD, and DepthFL, which do not leverage the server's computational capabilIties, we demonstrate only one performance metric achievable from the global model. In contrast, for AccSFL and FedSplitX, which utilize the server's computational power, we have presented two accuracies: one based on the client-side model and the other based on the combined performance of the client-side and server-side models in the full model configuration. For inference, FedSplitX, like DepthFL, uses the ensemble of all auxiliary networks in the model.

\subsection{Experimental Setup}

\textbf{Datasets and Models}~~~
To evaluate our proposed FedSplitX, we perform experiments with CFIAR-10, CIFAR-100 \cite{cifar10} datasets for image classification and the dataset is independent and identically distributed (IID) to the clients. For data augmentation, we applied random cropping, random horizontal flip, and normalization during data preprocessing. We utilized ResNet18, ResNet34, ResNet50, and ResNet101 \cite{he2016deep}. For all experiments in FedSplitX and baselines, three depth levels $d_k = 1,2,3, \;k \in [K]$ are considered for each model. The number of FLOPs and parameters for client-side based on depth-levels for each model type is summarized in the Table \ref{flops_params}. 16, 16, 18 clients at depth level 1, level 2 and level 3, respectively.

\textbf{Baselines}~~~
We compare our proposed algorithm FedSplitX with the following four baseline approaches: \lowercase\expandafter{\romannumeral1}) Exclusive learning (EXC), \lowercase\expandafter{\romannumeral2}) FjORD \cite{horvath2021fjord}, \lowercase\expandafter{\romannumeral3}) DepthFL \cite{kim2023depthfl}, \lowercase\expandafter{\romannumeral4}) Accelerated SFL (AccSFL) \cite{han2021accelerating}.
In EXC, only clients that can train the global model of each level will participate in the training, and incapable clients will not be able to participate. We considered the global model equivalent to client-side models of level 1, level 2, and level 3, respectively. For example, EXC of the global model equivalent to the client-side model of depth level 3, only 18 clients participate in the training. AccSFL is an homogeneous SFL framework which reduces communication costs by connecting an auxiliary network to the end of the client-side model. It considers a fixed partition point that only clients that meet the criteria can participate in the training, just as in EXC. FjORD and DepthFL refer to a width/depth pruning FL framework that address the problem of heterogeneous resource constrained by training sub-models which are extracted from global model, based on each client's individual capabilities. See Figure \ref{fig:structure} in Appendix for an illustration of the local model and global model used for each method.

\begin{table}[t]
\caption{Performance of \textbf{FedSplitX (ours)} and baselines with $M = 3$ depth-levels for CFIAR-10 dataset under IID settings. We report Top-1 classification accuracy (\%) of the client-side and full model performance per depth levels: (client-side model accuracy / full-model accuracy)}
\label{results}
\begin{center}
\resizebox*{0.7\textwidth}{!}{
\centerline{
\begin{tabular}{clcccclccc}
\toprule
\multirow{3}{*}{Model}    & \multirow{3}{*}{Method} & \multicolumn{3}{c}{Depth-level}                                                                                                                                                                       & \multirow{3}{*}{Model}        & \multirow{3}{*}{Method}                                                               & \multicolumn{3}{c}{Depth-level}  \\ \cmidrule{3-5} \cmidrule{8-10} 
                          &  & Level 1  & Level 2 & Level 3& &  & Level 1& Level 2 & Level 3                                                           \\ \midrule \midrule
\multirow{7}{*}{ResNet18} & EXC & 38.43  & 82.54  & 82.89                                                       
& \multirow{7}{*}{ResNet50}     & EXC& 38.42 & 84.99& 83.06                                \\ \cmidrule{2-5} \cmidrule{7-10} 
& FjORD & 63.30&78.08 & 78.12& & FjORD  & 17.7& 35.11& 35.10\\ \cmidrule{2-5} \cmidrule{7-10} 
& DepthFL & 40.50& 67.38& 74.44   & & DepthFL  & 37.89  & 70.08 & 75.42                                                     \\ \cmidrule{2-5} \cmidrule{7-10} 
& AccSFL  & 39.34 / 88.27&  82.07 / 84.68  &  83.58 / 83.74   &   & AccSFL  & 38.75 / 85.77 & 84.87 / 85.32 & 83.17 / 82.88   \\ \cmidrule{2-5} \cmidrule{7-10} 
& \textbf{FedSplitX}  & 51.81 / 80.91&  80.36 / 84.01&  83.81 / 84.4& 
& \textbf{FedSplitX} & 50.75 / 79.34& 84.7 / 85.04&  85.69 / 85.7\\ \midrule

\multirow{7}{*}{ResNet34} & EXC & 38.43   & 86.63 & 83.92 & \multirow{7}{*}{ResNet101}    & EXC & 38.4& 86.2  & 81.91                                                             \\ \cmidrule{2-5} \cmidrule{7-10} 
& FjORD &55.86 &74.71 &74.69 & & FjORD&18.53 &35.13 &35.13 \\ \cmidrule{2-5} \cmidrule{7-10} 
& DepthFL & 40.04 & 77.04  & 82.24  &                                                       & DepthFL  & 31.72& 69.26 & 71.95                                                          \\ \cmidrule{2-5} \cmidrule{7-10}
& AccSFL  & 39.52 / 88.03& 86.44 / 86.42& 84.08 / 83.89    &                               
& AccSFL &  38.98 / 83.39 & 86.15 / 86.23 &  81.27 / 80.89                                   \\ \cmidrule{2-5} \cmidrule{7-10} 
&  \textbf{FedSplitX}              & 51.74 / 82.73& 85.36 / 86.02 & 86.48 / 86.53 &                               &  \textbf{FedSplitX}                                                                            & 51.19 / 78.3 & 85.06 / 85.64 & 86.13 / 85.98 \\ 
                          \bottomrule
\end{tabular}
}
}
\end{center}
\end{table}

\subsection{Comparison with Baselines} 
We present that FedSplitX enables the clients to train large models by utilizing the server's capabilities.
In cases like EXC at depth-level 1, the small size of the global model enables the participation of all clients in training process. However, as shown in Table \ref{results}, the performance of global model in EXC at depth-level 1 is significantly lower than the full model accuracy achieved by FedSplitX due to the reduced global model size. 
When we compare the performance of FedSplitX's client-side model and EXC's global model, we can observe that performance of FedSplitX outperforms EXC at both level 1 and level 3. While the global model in EXC and the client-side model in FedSplitX are equivalent at each depth-level, the performance of FedSplitX which uses server capabilities, proved to be better. 
FjORD and DepthFL extract sub-models from the global model to meet the capabilities of clients, enabling clients with poor capabilities to participate in the training. However, we considered a global model size for DepthFL and FjORD equivalent to the client-side model at depth-level 3, which is smaller than the full model trained by FedSplitX. Consequently, training with FedSplitX results in better performance than both baseline methods. In particular, FjORD adjusts its dropout ratio to match FedSplitX's FLOPs, leading to excessive model narrowing and further performance degradation. These outcomes highlight the advantage of partitioning the model into client-side and server-side components, enabling learning beyond individual client capabilities and contributing to performance improvement.

Another significant advantage of FedSplitX is ability to involve all clients in the training process, regardless of their individual computing capabilities. In the EXC approach, as the desired global model size increases, clients with limited capabilities were unable to participate in the training. However, FedSplitX offers the flexibility of partition points based on client abilities, enabling the customization of client-side models to meet their capabilities. As a result, clients with poor resources can participate in the learning process.
AccSFL and FedSplitX both leverage server capabilities for training, but ACCSFL differs by offering a fixed partition point. The size of the client-side model may heterogeneous depending on the location of the fixed partition point, potentially limiting the number of participating clients in the training process. 
Comparing the performance of the client-side and full models between AccSFL and FedSplitX at depth-level 3 demonstrates that FedSplitX achieves better performance as it involves a larger number of participating clients compared to AccSFL.

In AccSFL, the fixed partition point located at the front results in the client-side model becoming smaller, consequently leading to an improvement in the performance of the full model as the number of participating clients in training increases. We can observe that the improved performance of AccSFL is better than full model FedSplitX at depth-level 1. However, the fixed partition point at the front of the model leads to an increase in the size of the server-side model, subsequently placing a higher computational load on the server. FedSplitX is an algorithm that provides multiple partition points based on client power, so that more powerful clients can fully utilize their power, thus reducing the amount of computation loads on the server. The Table \ref{acc_fedsplitx} shows that the server computes several times more FLOPs with AccSFL than with FedSplitX.

\begin{table}[h]
  \caption{Comparison computation cost between AccSFL(depth-level 1) and FedSplitX. We report the number of FLOPs required to the server. For FedSplitX, we have averaged the server-side FLOPs over three depth levels. }
  \label{acc_fedsplitx}
  \centering
  \resizebox*{!}{0.105\textheight}{
  \begin{tabular}{cccc}
    \toprule
    Model  & AccSFL (depth-level 1) &FedSplitX \\ \midrule
    {ResNet18} & 138.4M&   102.1M \\
    \midrule
    {ResNet34} &  289.5M &  205.6M   \\
    \midrule
    {ResNet50} & 307.9M&    218.8M    \\
    \midrule
    {ResNet101} & 628.5M&  456.3M \\
    \bottomrule
  \end{tabular}
  }
\end{table}

\section{Conclusion}
\label{others}
In this work, we proposed \textit{FedSplitX}, a novel federated split learning algorithm designed to tackle the challenges of federated learning in heterogeneous and resource-constrained client systems. 
FedSplitX effectively splits a large model into two components, harnessing the server's capabilities to facilitate training of larger models than individual clients can handle, consequently leading to enhanced performance. Furthermore, providing multiple partition points based on client abilities ensures participation of all clients, utilizes their full capabilities, and reduces unnecessary server workload. Our experimental results demonstrate that FedSplitX outperforms the baseline method that relies solely on client capabilities without utilizing server resources. Furthermore, by enabling clients with diverse resources to participate in the learning process, FedSplitX maintains fairness and enhances learning performance. As future work, we plan to analyze how performance varies with partition point, and to study what the trends are in larger models.

\medskip
\newpage

\bibliography{JYref}
\bibliographystyle{ieee_fullname}

\newpage
\appendix

{\Large \textbf{Supplementary Material}}

\section{Split Model with Multiple Partition Points}

\subsection{Partition Points based on Depth-Level}
For all experiments, we consider $M = 3$ depth-levels. The models are split at the following partition points. The Table \ref{flops_params} summarizes the number of FLOPs and parameters per model for the client-side model split according to the split points below. 

\textbf{ResNet18}~~~ the partition point for depth-level $1$ is between 2D MaxPool layer (after convolutional layer - 2D BatchNormalization layer) and the first ResBlock. The partition point for depth-level $2$ is between first ResBlock and second ResBlock. The last partition point is between second ResBlock and third ResBlock.

\textbf{ResNet34, ResNet50}~~~ the partition point for depth-level $1$ is between 2D MaxPool layer (after convolutional layer - 2D BatchNormalization layer) and the first ResBlock. The partition point for depth-level $2$ is between second layer of second ResBlock and third layer of second ResBlock. The last partition point is between second layer and third layer of third ResBlock.

\textbf{ResNet101}~~~ the partition point for depth-level $1$ is between 2D MaxPool layer (after convolutional layer - 2D BatchNormalization layer) and the first ResBlock. The partition point for depth-level $2$ is between first layer and second layer of third ResBlock. The last partition point is between eleventh layer and twelfth layer of third ResBlock.

\begin{table}[htp]
  \caption{Number of FLOPs (left) and parameters (right) of client-side model according depth-level}
  \label{flops_params}
  \centering
  \resizebox*{!}{0.11\textheight}{
  \begin{tabular}{cccc}
    \toprule
    \multirow{3}{*}{Model}    & \multicolumn{3}{c}{Depth level}        \\ \cmidrule(l){2-4} 
                              & Level 1   & Level 2 & Level 3 \\ \midrule \midrule
    {ResNet18} & 1.77M (1\%) &  39.53M (28\%) &  73.10M (52\%) \\
    \midrule
    {ResNet34} &  1.77M (0.6\%) &  91.96M (31\%) & 163.3M(56\%)\\
    \midrule
    {ResNet50} & 1.77M(0.6\%) &  96.74M (31\%) &  177.5M (57\%)\\
    \midrule
    {ResNet101} & 1.77M (0.3\%)&   172M (27\%)&  350.2M (56\%)\\
    \bottomrule
  \end{tabular}
  }
    \resizebox*{!}{0.108\textheight}{
    \begin{tabular}{cccc}
    \toprule
    \multirow{3}{*}{Model}    & \multicolumn{3}{c}{Depth level}        \\ \cmidrule(l){2-4} 
                              & Level 1   & Level 2 & Level 3 \\ \midrule \midrule
    {ResNet18} & 4.3K (0.04\%) &  154.7K (1.38\%) &  689.4K (6.17\%) \\
    \midrule
    {ResNet34} &  4.3K (0.02\%) &  763.4K (3.59\%) & 3.488M (16.23\%)\\
    \midrule
    {ResNet50} & 4.3K (0.02\%) &  1.05M (4.46\%) &  4.77M (20.28\%)\\
    \midrule
    {ResNet101} & 4.3K (0.01\%)&   3.65M (8.59\%)&  14.82M (34.87\%)\\
    \bottomrule
  \end{tabular}
  } 
\end{table}

\subsection{Local Model in Baselines}

\begin{figure}[t]
    \centering
    \begin{subfigure}{0.75\textwidth}
        \centering
         \noindent
        \makebox[\textwidth]{
        \includegraphics[width=0.8\textwidth]{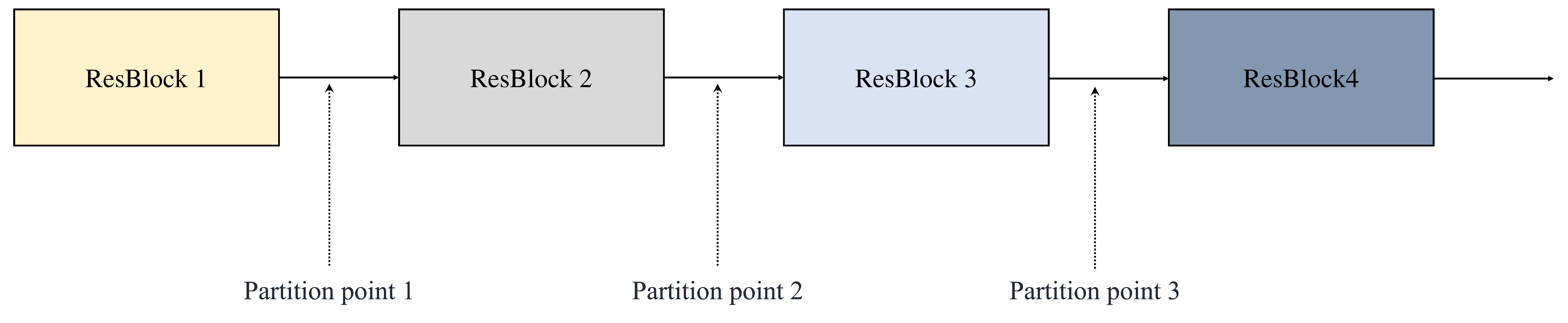}
        }
                            \caption{Structure of ResNet (full model)}
        \label{fig:orig}
    \end{subfigure}
    
    \begin{subfigure}{0.75\textwidth}
        \centering
         \noindent
        \makebox[\textwidth]{
        \includegraphics[width=1\textwidth]{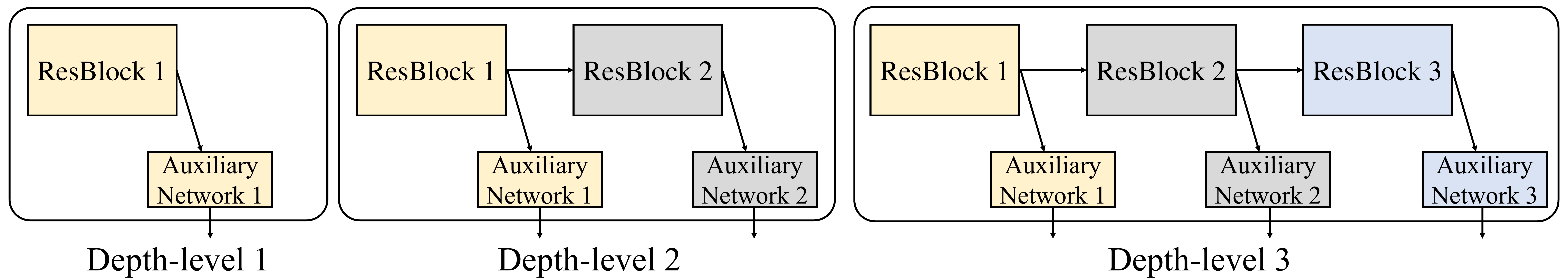}
        }
            \caption{Structure of the local model trained from the EXC}
        \label{fig:exc}
    \end{subfigure}
        \begin{subfigure}{0.7\textwidth}
        \centering
         \noindent
        \makebox[\textwidth]{
        \includegraphics[width=1\textwidth]{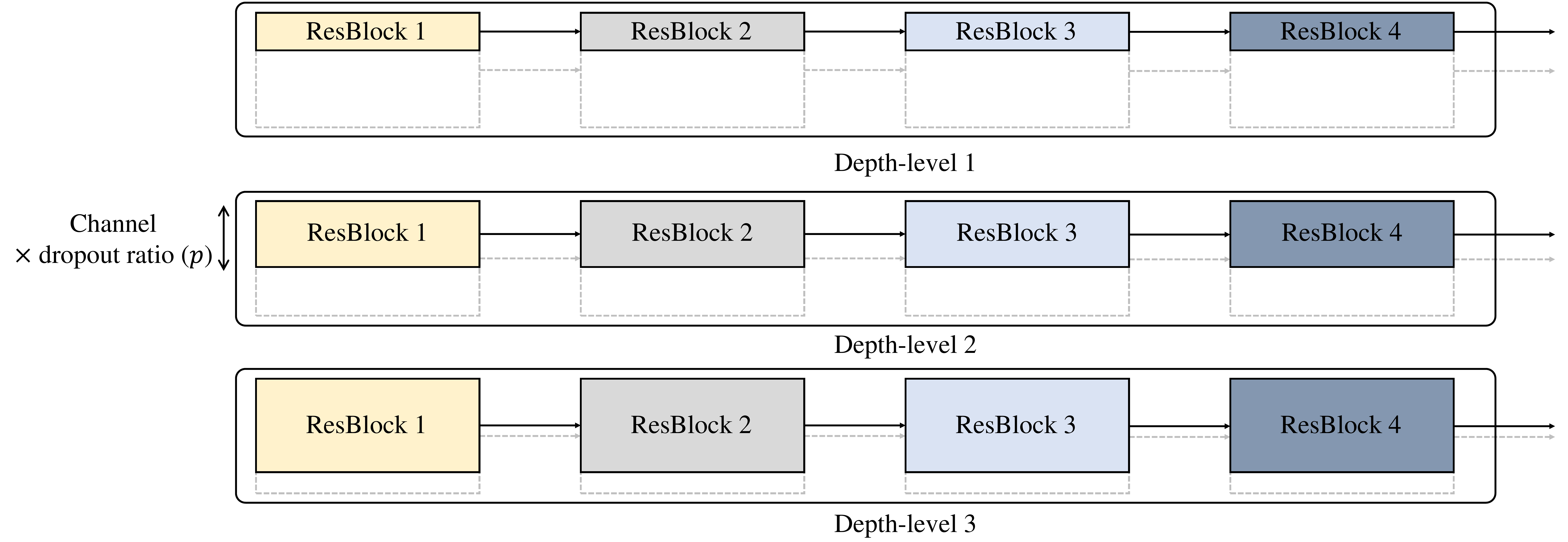}
        }
            \caption{Structure of the local model trained from the FjORD}
        \label{fig:fjord}
    \end{subfigure}
     \begin{subfigure}{0.7\textwidth}
        \centering
         \noindent
        \makebox[\textwidth]{
        \includegraphics[width=1\textwidth]{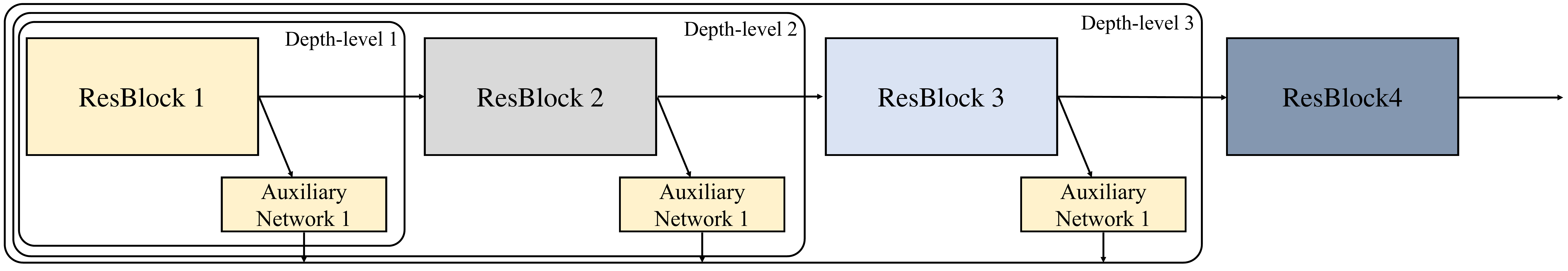}
        }
            \caption{Structure of the local model trained from the DetphFL}
        \label{fig:depth}
    \end{subfigure}
        \begin{subfigure}{0.75\textwidth}
        \centering
         \noindent
        \makebox[\textwidth]{
        \includegraphics[width=1\textwidth]{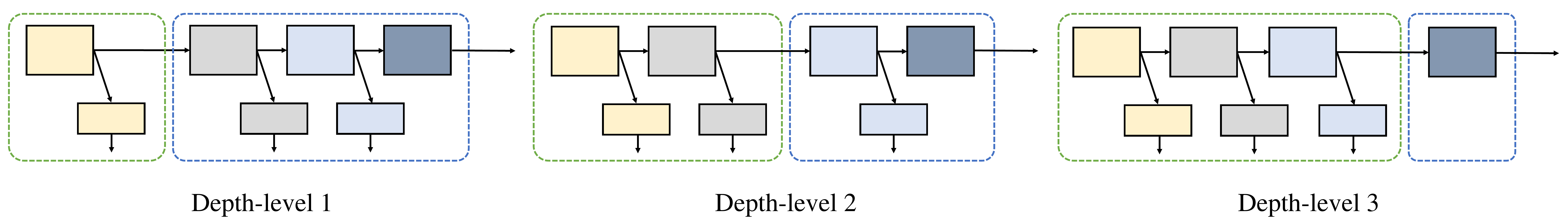}
        }
            \caption{Structure of the local model trained from the FedSplitX and AccSFL}
        \label{fig:fedsplitx}
    \end{subfigure}

    \caption{Structure of the local model trained from the FedSplitX and baselines based on the depth-level}
    \label{fig:structure}
\end{figure}

The structure of the local model trained in the baseline, based on the depth-level, is shown in Figure \ref{fig:structure}. For the client system we considered, we assume that the largest model that the most capable depth-level 3 client can train corresponds to the client-side model at depth-level 3.  In other words, the global model for baselines that do not utilize the server's capabilities is smaller than the full model. Consequently, for a baseline approach like EXC(Figure \ref{fig:exc})), DepthFL(Figure \ref{fig:depth}), and FjORD(Figure \ref{fig:fjord}) that doesn't leverage the server's capabilities, the maximum trainable size of the global model does not exceed that of the full model of FedSplitX(Figure \ref{fig:fedsplitx}).
Unlike FedSplitX and DepthFL, which scale models based on depth, FjORD scales the width while extracting sub-models that fit the client's capabilities. Therefore, it is difficult to have a local model that is exactly the same as the local model adjusted by depth. We conducted experiments using dropout rates that yielded equivalent FLOPs to those associated with the depth-level analyzed in FedSplitX, as indicated in the Table \ref{dropout}.

\begin{table}[htp]
  \caption{Dropout rate based on the depth-level using FjORD.}
  \label{dropout}
  \centering
  \resizebox*{!}{0.11\textheight}{
  \begin{tabular}{cccc}
    \toprule
    \multirow{3}{*}{Model}    & \multicolumn{3}{c}{Depth level}        \\ \cmidrule(l){2-4} 
                              & Level 1   & Level 2 & Level 3 \\ \midrule \midrule
    {ResNet18} & 0.1 &  0.53 &  0.72 \\
    \midrule
    {ResNet34} &  0.07 &  0.56 & 0.74\\
    \midrule
    {ResNet50} &  0.07 &  0.56 & 0.76\\
    \midrule
    {ResNet101} & 0.05 &  0.52 & 0.74\\
    \bottomrule
  \end{tabular}
  }
\end{table}

\newpage

\section{Additional Experiments}
\subsection{Other Dataset}
We evaluate the performance for other dataset such as CIFAR-100. The results are presented in Table \ref{result_cifar100}.

\begin{table}[b]
\caption{Performance of \textbf{FedSplitX (ours)} and baselines with $M = 3$ depth-levels for \textbf{CFIAR-100} dataset under IID settings. We report Top-1 classification accuracy (\%) of the performance per depth levels}
\label{result_cifar100}
\begin{center}
\resizebox*{0.7\textwidth}{!}{
\centerline{
\begin{tabular}{clcccclccc}
\toprule
\multirow{3}{*}{Model}    & \multirow{3}{*}{Method} & \multicolumn{3}{c}{Depth-level}                                                                                                                                                                       & \multirow{3}{*}{Model}        & \multirow{3}{*}{Method}                                                               & \multicolumn{3}{c}{Depth-level}  \\ \cmidrule{3-5} \cmidrule{8-10} 
                          &  & Level 1  & Level 2 & Level 3& &  & Level 1& Level 2 & Level 3                                                           \\ \midrule \midrule
\multirow{7}{*}{ResNet18} & EXC & 14.37  & 48.63  & 52.18                                                    
& \multirow{7}{*}{ResNet50}     & EXC & 14.37 & 56.97 & 48.78                              \\ \cmidrule{2-5} \cmidrule{7-10} 

& FjORD & 22.51& 42.31& 42.43& & FjORD  & 3.78&12.74 & 12.75\\ \cmidrule{2-5} \cmidrule{7-10} 
& DepthFL & 13.93& 37.78& 46.71   & & DepthFL  & 14.37 & 49.50& 53.83                                                     \\ \cmidrule{2-5} \cmidrule{7-10} 

& AccSFL  & 14.34 / 61.32&  48.05 / 53.91  &  52.38 / 51.00   &   & AccSFL  & 14.44 / 57.2 & 57.07 / 57.55 & 48.66 / 46.83   \\ \cmidrule{2-5} \cmidrule{7-10} 
& \textbf{FedSplitX}  & 20.16 / 51.02&  43.79 / 54.23&  50.83 / 53.58& 
& \textbf{FedSplitX} & 30.26 / 41.44& 56.86 / 57.45&  57.86 / 57.16\\ \midrule

\multirow{7}{*}{ResNet34} & EXC &  14.38 & 59.25 & 55.34 & \multirow{7}{*}{ResNet101}    & EXC & 14.37& 58.45  &                                                              49.86\\ \cmidrule{2-5} \cmidrule{7-10} 
& FjORD & 15.47&38.53 &38.56 & & FjORD& 3.55&12.98 &12.99 \\ \cmidrule{2-5} \cmidrule{7-10} 
& DepthFL & 14.67 & 54.03  & 58.13  &  & DepthFL  & 13.77& 53.13 & 55.35                                                          \\ \cmidrule{2-5} \cmidrule{7-10}

& AccSFL  &14.43 / 58.63 & 58.71 / 58.15& 51.15 / 50.16   &                               
& AccSFL &  14.46 / 55.49 & 58.7 / 56.98 &  46.56 / 44.41                                   \\ \cmidrule{2-5} \cmidrule{7-10} 
&  \textbf{FedSplitX}              & 30.27 / 42.96& 56.72 / 57.12 & 58.34 / 57.95 &                               
&  \textbf{FedSplitX} &19.08/ 49.97& 56.99 / 56.95 & 57.67 / 55.93 \\ 
                          \bottomrule
\end{tabular}
}
}
\end{center}
\end{table}

\subsection{Ablation Study}
\textbf{Effect of collaborative loss}

In the FedSplitX framework we propose, we utilize auxiliary networks at each partition point, enabling clients to train independently without the need for gradient communication from the server. Additionally, we leverage the outputs of these auxiliary networks to compute a collaborative loss, which enhances the effectiveness of sub-model learning. To assess whether collaborative loss contributes to performance improvement and reduces performance heterogeneity among models, we conducted experiments by removing all auxiliary networks from FedSplitX, except for the auxiliary network attached to the end of the client-side model. We then compared the performance of this modified version, called FedSplitX w/o auxnet, with the original FedSplitX. Refer the Table \ref{ablation_cifar10},\ref{ablation_cifar100} performance of FedSplitX, which updates the model by calculating the collaborative loss from the auxiliary network is much better than FedSplitX w/o auxnet. By utilizing the logits at every partition point, we observed a reduction in performance heterogeneity between models. Interestingly, even in cases where the performance gap increased, we could see an improvement in overall performance.
\begin{table}[t]
\caption{Ablation study by FedSplitX w/o auxnet and FedSplitX (original) with $M=3$ depth-levelsfor CIFAR-10 dataset under IID settings. We report Top-1 classification accuracy (\%) according to the depth-level.}
\label{ablation_cifar10}
\centering
\resizebox*{0.85\textwidth}{!}{
\begin{tabular}{@{}clcccc@{}}
    \toprule
    \multirow{3}{*}{Model}    & \multirow{3}{*}{Method} & \multicolumn{3}{c}{Depth level}  & \multirow{3}{*}{ Standard Deviation}    \\ \cmidrule(l){3-5} 
                              & & Level 1   & Level 2 & Level 3&  \\ \midrule \midrule
    \multirow{3}{*}{ResNet18}   
                              &  FedSplitX w/o auxnet  &  81.53 &  84.28 & 82.49 & 1.40\\ \cmidrule(l){2-6} 
                              &  \textbf{FedSplitX (ours)} & 80.91& 84.01	& 84.4 & 1.91\\ \midrule
    \multirow{3}{*}{ResNet34}
                              &  FedSplitX w/o auxnet  &    75.28&  83.92& 84.99 & 5.32\\ \cmidrule(l){2-6} 
                              &  \textbf{FedSplitX (ours)} & 82.73& 86.0& 86.53 & 2.06\\ \midrule
    \multirow{3}{*}{ResNet50}
                              &  FedSplitX w/o auxnet  &  70.5	&81.59&	82.69 & 6.742 \\ \cmidrule(l){2-6} 
                              &  \textbf{FedSplitX (ours)} & 79.34	&85.04	&85.7 & 3.50 \\ \midrule
    \multirow{3}{*}{ResNet101}
                              &  FedSplitX w/o auxnet  &  71.99	& 82.61 & 	77.61 & 5.31\\ \cmidrule(l){2-6} 
                              &  \textbf{FedSplitX (ours)} & 78.& 85.64	& 85.98 & 4.34  \\\midrule
\end{tabular}
}
\end{table}
\begin{table}[t]
\caption{Ablation study by FedSplitX w/o auxnet and FedSplitX (original) with $M=3$ depth-levelsfor CIFAR-100 dataset under IID settings. We report Top-1 classification accuracy (\%) according to the depth-level.}
\label{ablation_cifar100}
\centering
\resizebox*{0.85\textwidth}{!}{
\begin{tabular}{@{}clcccc@{}}
    \toprule
    \multirow{3}{*}{Model}    & \multirow{3}{*}{Method} & \multicolumn{3}{c}{Depth level}  & \multirow{3}{*}{ Standard Deviation}    \\ \cmidrule(l){3-5} 
                              & & Level 1   & Level 2 & Level 3&  \\ \midrule \midrule
    \multirow{3}{*}{ResNet18}   
                              &  FedSplitX w/o auxnet  & 52.58&	55.05	& 45.94 &4.71 \\ \cmidrule(l){2-6} 
                              &  \textbf{FedSplitX (ours)} &51.02&	54.23&53.58& 1.7\\ \midrule
    \multirow{3}{*}{ResNet34}
                              &  FedSplitX w/o auxnet  &    48.03&	52.61&	48.71 &2.47\\ \cmidrule(l){2-6} 
                              &  \textbf{FedSplitX (ours)} & 42.96& 57.12&	57.95 &8.43\\ \midrule
    \multirow{3}{*}{ResNet50}
                              &  FedSplitX w/o auxnet  &  45.63&	51.09&	48.40&2.73 \\ \cmidrule(l){2-6} 
                              &  \textbf{FedSplitX (ours)} & 41.44&	57.45&	57.163&9.16\\ \midrule
    \multirow{3}{*}{ResNet101}
                              &  FedSplitX w/o auxnet  &  41.66&	47.58&	39.68& 4.11\\ \cmidrule(l){2-6} 
                              &  \textbf{FedSplitX (ours)} & 50.00&	56.95&	55.93& 3.77 \\ \midrule
\end{tabular}
}
\end{table}

\end{document}